\def\year{2018}\relax
\documentclass[letterpaper]{article} 
\usepackage{aaai18}  
\usepackage{times}  
\usepackage{helvet}  
\usepackage{courier}  
\usepackage{url}  
\usepackage{graphicx}  
\usepackage{times} 
\usepackage{helvet}
\usepackage{courier} 
\usepackage{psfrag} 
\usepackage{fleqn}
\usepackage{latexsym} 
\usepackage{color} 
\usepackage{amsmath}
\usepackage{amsfonts} 
\usepackage{amssymb} 
\usepackage[T1]{fontenc}
\usepackage{plain}
\usepackage{url}
\usepackage{caption}
\usepackage{float}
\usepackage{amsthm}

\newcommand\finalitem{\item[\theenumi']}

\begin{document}

\newcommand{\Omit}[1]{}
\newcommand{\denselist}{\itemsep -1pt\partopsep 0pt}
\newcommand{\tup}[1]{\langle #1 \rangle}
\newcommand{\pair}[1]{\langle #1 \rangle}
\newtheorem{definition}{Definition}
\newtheorem{definitionandtheorem}[definition]{Definition and Theorem}
\newtheorem{theorem}[definition]{Theorem}
\newtheorem{lemma}[definition]{Lemma}
\newtheorem{proposition}[definition]{Proposition}
\newtheorem{corollary}[definition]{Corollary}
\newenvironment{sketch}{\noindent \emph{Proof sketch:}}{\hfill$\square$}

\title{Compact Policies for Fully-Observable Non-Deterministic Planning as SAT}

 \author{Tomas Geffner \\
  University of Massachusetts \\
   Amherst, USA \\
     {\normalsize\url{tomas.geffner@gmail.com}}
    \And
    Hector Geffner \\
    ICREA \&  Universitat Pompeu Fabra \\
    Barcelona, SPAIN \\
    {\normalsize\url{hector.geffner@upf.edu}}}

\date{}
\maketitle

\begin{abstract}
Fully observable non-deterministic (FOND) planning is becoming increasingly important
as an approach for computing proper policies in probabilistic planning,  extended temporal plans  in LTL planning,
and  general  plans  in generalized planning. In this work, we introduce a SAT encoding for FOND planning
that is   compact and can produce compact strong cyclic policies.   Simple variations of the encodings are also introduced  for strong planning
and for what we call, dual FOND planning, where some non-deterministic actions are  assumed to be fair (e.g., probabilistic)
and others  unfair (e.g., adversarial). The resulting FOND planners are compared empirically with existing planners over
existing and new benchmarks. The notion of ``probabilistic interesting problems'' is also revisited to  yield
a more comprehensive picture of the strengths and limitations of current FOND planners and the proposed SAT approach.
\end{abstract}

\section{Introduction}

Planning is the model-based approach to autonomous behavior. A planner produces
a plan for a given goal and initial situation using a model of  the actions and sensors.
In fully-observable non-deterministic (FOND) planning, actions may have non-deterministic effect and
states are assumed to be  fully observable \cite{cimatti:three-models}. FOND planning is closely related to probabilistic
planning in Markov Decision Processes (MDPs), except that uncertainty about successor
states is represented by sets rather than  probabilities. However, the policies that achieve the goal 
with probability 1, the so-called proper policies \cite{bertsekas:neuro},   correspond exactly to the
strong cyclic policies of the associated FOND model \cite{strong-cyclic}  where the possible transitions
are those with non-zero probabilities \cite{geffner:book}.

FOND planning has become increasingly important as a way of  solving other types of problems.
In generalized planning, one is interested in plans that provide the solution to multiple, and even,
infinite collection of instances  \cite{srivastava:generalized,hu:generalized}.
For example, the  policy ``if end not visible, then move''  can take an agent to the end of a $1 \!\times\! n$
grid regardless of the initial location of the agent and  the value of  $n$.  In many cases general   policies  can be
obtained effectively from suitable FOND abstractions \cite{srivastava:aaai2011,bonet:ijcai2015}.  For example, the policy above
can be obtained from a  FOND abstraction over a $1 \!\times\! 2$ grid where the ``move'' actions become
non-deterministic and can leave the agent in the same cell. The non-determinism is a result of the abstraction
\cite{bonet:ijcai2017}. Planning for  extended temporal (LTL) goals like ``forever, visit each of the rooms eventually''  that
require  ``loopy'' plans have  also been reduced to FOND planning in many cases \cite{camacho:ltl}.
In such a  case, the non-determinism  enters  as a  device  for obtaining infinite executions.
For example, the extended temporal goal ``forever eventually $p$'' can be reduced to reaching the dummy goal 
$d$ once  the deterministic outcomes  $E$ containing  $p$  are replaced by the  non-deterministic outcomes
$oneof(E,d)$ \cite{nir:ijcai2013}.

In spite of the increasing importance of FOND planning, research on effective computational approaches
has slowed down in recent years. There are indeed  OBDD-based  planners like MBP and Gamer \cite{cimatti:three-models,gamer},
planners relying on explicit AND/OR graph search like MyND and Grendel \cite{mynd,grendel},  and
planners that rely on classical algorithms  like  NDP, FIP, and PRP \cite{nd,fip,prp}, yet recent
ideas,  formulations, and new benchmarks have been scarce. This may have to do with the fact one of these planners,
namely PRP, does incredibly well on the existing  FOND benchmarks. This is  somewhat surprising though,
given that non-determinism plays a passive role in the search for plans in PRP which  is based
on the computation of classical plans using the deterministic relaxation \cite{ff-replan}.

The goals of this work are twofold. On the one hand, we  want to improve the analysis
of the computational challenges in FOND planning  by  appealing to \emph{three dimensions
of analysis: problem size, policy size, and robust non-determinism.} For this last dimension
we provide a precise measure that refines the notion of  ``probabilistic interesting problems''
introduced for distinguishing  challenging forms of non-determinism from trivial ones 
\cite{pip}. Non-determinism is trivial when there is no need to take non-determinism into
account when reasoning about the future. This is the case when the risk involved is minimal.
In \emph{on-line} FOND  planning, the risk is the probability of not reaching the goal.
This is  the type of risk that Little and Thiebaux had in mind when they introduced
the notion of probabilistic interesting problems. On the other hand, in \emph{off-line}
FOND planning, any complete algorithm will reach the goal with probability 1 in
solvable problems. The ``risk'' in such a case lies in  the \emph{computational cost} of producing
one such solution. We will see that this computational cost for FOND planners based on classical planning 
can be estimated and used to analyze current  benchmarks and to introduce new ones.

The second goal of the paper is to introduce a new approach to FOND planning based on SAT.
The potential advantage of SAT approaches to FOND planning is that while classical replanners
reason about the possible executions of the plan one by one, the SAT approach performs inference
about all branching  executions in parallel (interleaved). Moreover, while  previous SAT approaches
to FOND planning rely on CNF encodings where there is a propositional symbol for each possible
state \cite{chitta:fond,chatterjee:sat}, we develop a compact encoding  that  can produce  compact
policies  too. That is, the size of the encodings does not grow with the number of states in the problem,
and the size of the resulting  policies does not necessarily grow with the number of  states that are reachable with
the policy. Simple variations of the encoding are introduced  for  \emph{strong planning}  and for what we call
\emph{dual FOND planning,}  where some non-deterministic actions are  assumed to be fair (e.g., probabilistic) and others
unfair (e.g., adversarial). The resulting SAT-based  FOND planner is   compared empirically with Gamer, MyND,
and PRP.

The paper is organized as follows. We review FOND planning, classical approaches, and the challenge of non-determinism.
We introduce then the new SAT approach, the  formal properties, optimizations that preserve these properties, 
and the  evaluation. We look then at the variations required for strong and dual FOND planning, and draw  final conclusions.

\section{FOND Planning}

A FOND model is a tuple $M = \tup{S,s_0,S_G,Act,A,F}$ where $S$ is a finite set of states, $s_0 \in S$ is the initial state,
$S_G \subseteq S$ is a non-empty set of goal states, $Act$ is a set of actions, $A(s) \subseteq Act$ is the set of actions applicable
in the state $s$, and $F(a,s)$ for $a \in A(s)$ represents the non-empty set of successor states that follow action $a$ in state $s$.
A FOND problem $P$ is a compact description of a FOND model $M(P)$ in terms of a finite set  of atoms, so that the
the states $s$ in $M(P)$ correspond to truth valuations over the atoms,  represented by the set of atoms that are true.
The standard syntax for FOND problems is a simple extension of the STRIPS syntax for classical planning.
A FOND problem is a tuple $P = \tup{At,I,Act,G}$ where $At$ is a set of atoms, $I \subseteq At$ is the
set of atoms true in the initial state $s_0$, $G$ is the set of goal atoms, and $Act$ is a set of actions
with atomic preconditions and effects.  If $E_i$ represents  the set of positive and negative effects of an  action in the classical setting,
action effects in FOND planning can be deterministic of  the form $E_i$, or non-deterministic of the form $oneof(E_1,\ldots,E_n)$.
Alternatively, a \emph{non-deterministic} action $a$ with effect $oneof(E_1,\ldots,E_n)$ can be regarded as a set of \emph{deterministic}
actions $b_1$, \ldots, $b_n$ with effects $E_1$, \ldots, $E_n$ respectively, written as $a=\{b_1, \ldots, b_n\}$, all  sharing the same preconditions
of $a$. The application of $a$ results in the application of one of the actions $b_i$ chosen non-deterministically.

A policy $\pi$ for a FOND problem $P$ is a partial function mapping \emph{non-goal} states $s$ into actions $a \in A(s)$.
A policy $\pi$ for $P$ induces   state trajectories $s_0, \ldots,s_n$ where $s_{i+1} \in F(a_i,s_i)$ and  $a_i = \pi(s_i)$
for $i=0, \ldots, n-1$. A  trajectory $s_0, \ldots,s_n$ induced by $\pi$ is \emph{complete}
if $s_n$ is the first state in the sequence such that $s_n$ is a goal state or  $\pi(s_n) = \bot$,
or if the trajectory is not finite. Similarly, the trajectory induced by the policy $\pi$ is \emph{fair} if it is finite,
or if infinite occurrences of states $s$  in the trajectory with $\pi(s)$ being the action   $a=\{b_1, \ldots, b_m\}$ are
followed an infinite number of times by the state $s_i$ that  results from $s$ and $b_i$, for each $i=1, \ldots, m$.
A policy $\pi$ is a \emph{strong solution} for $P$ if  the complete state trajectories induced by $\pi$
are all  goal reaching, and it is a \emph{strong cyclic solution} for $P$ if  the complete state trajectories induced by $\pi$ that
are \emph{fair} are all goal reaching.
Strong and strong cyclic solutions are also called strong and strong cyclic policies for $P$
respectively.

The methods for computing strong and strong cyclic solutions to FOND problems have been mostly based on 
OBDDs  \cite{cimatti:three-models,gamer},  explicit forms of  AND/OR search \cite{mynd,grendel},
and  classical  planners \cite{nd,fip,prp}. Some of the planners compute \emph{compact}
policies in the sense that the size of the  policies, measured by their representation,
can be  exponentially smaller than  the number of  states reachable with the policy. 
This is crucial in some  benchmark domains where the number of states reachable in the
solution is exponential in the problem size.

\section{Classical Replanning for FOND Planning}

The  FOND planners that  scale up   best  are   built on top of classical planners.
These planners, that we  call \emph{classical replanners}, all follow a loop where many classical plans are
computed until the set of classical plans forms a strong cyclic policy. In this loop, \emph{non-determinism  plays a passive role:}
it is not taken  into account for computing the classical plans but for determining which plans are still missing if any.
The good  performance of  these  planners is a result of  the  robustness and scalability of classical planners
and  the types of  benchmarks considered so far. 
We describe classical replanners for FOND first, and  turn then to three dimensions for analyzing challenges and benchmarks.

The (all-outcome) \emph{deterministic relaxation}  of a FOND problem $P$ is obtained by replacing each
non-deterministic action $a=\{b_1, \ldots, b_n\}$ by the set of deterministic actions  $b_i \in a$. 
A \emph{weak plan} in $P$ refers to a \emph{classical plan}  for the deterministic relaxation of $P$.

For a given FOND problem $P$, complete classical replanners yield  strong cyclic policies that solve $P$ by
computing a partial function $\rho$ mapping non-goal states $s$ into classical plans $\rho(s)$ for the deterministic
relaxation of $P$ with initial  state $s$. We write  $\rho(s)=b,\rho'$ to denote a plan for $s$ in the relaxation
that starts with the action $b$  followed by the action  sequence $\rho'$.
The following conditions ensure   that  the partial function $\rho$ encodes
a strong cyclic policy for $P$ \cite{geffner:book}:

\begin{enumerate} \denselist
\item Init: $\rho(s_0) \not= \bot$, 
\item Consistency: If $\rho(s)=b,\rho'$ and $s'=f(b,s)$,  $\rho(s')= \rho'$,
\item Closure: If $\rho(s)=b,\rho'$, $\forall \, s' \in F(b,s)$,  $\rho(s') \not= \bot$.
\end{enumerate}

In these conditions, $f(b,s)$ denotes the single next state for  actions $b$ in the relaxation,
while $F(b,s)$ denotes the set of possible successor states for actions in the original problem $P$, 
with $F(b,s)$  thus set to $F(a,s)$ when $b  \in a$. 

A partial function $\rho$ that complies with  conditions 1--3 encodes a strong cyclic solution $\pi$ to $P$.
If $\rho(s)$ is the plan $b_1, \ldots, b_n$ in the relaxation then  $\pi(s)=b_1$ if $b_1$ is a deterministic
action in $P$, else $\pi(s)=a$ for  $b_1 \in a$. Any strong cyclic plan for $P$ can be expressed
as a partial mapping of states into plans for the relaxation. The different classical replanners construct
the function $\rho$  in different ways usually starting with a first plan  $\rho(s_0)$,  enforcing then consistency and closure.
In a problem with no deadend states, the process finishes monotonically in a number of iterations and classical planner calls that is bounded by the number
of states that are reachable with the policy.  PRP uses  regression to reduce this number, 
resulting in  policies that map partial states into actions and may have an exponentially smaller size.
In the presence of deadends, the computation in PRP is similar but the process is restarted from scratch
with more action-state pairs excluded each time that  the classical planner fails to find a plan and 
close the function $\rho$. An additional component of PRP is an algorithm for inferring  and generalizing 
deadends that in certain cases can exclude many weak plans from consideration in one shot. 

\section{Challenges in FOND Planning}



The challenges in (exact) FOND planning have to do with three dimensions:
problem size,  policy size, and robust non-determinism.

\medskip

\noindent \textbf{Problem Size.} 
The size of the state space $M(P)$ for a FOND problem $P$
is exponential in the number of problem atoms. This is like in classical planning.
Approaches relying on classical planners, in particular PRP, appear to be the ones
that scale  up best. This, by itself, however, is not surprising given that classical planning problems
are (deterministic) FOND problems and   FOND  approaches that do not rely on classical planners
won't be as competitive. The trivial conclusion is that problem size alone is likely to  exclude
non-classical approaches to FOND planning in certain classes of problems.

\medskip

\noindent\textbf{Policy Size.}  Many FOND problems have solutions  of exponential size. This situation is uncommon in classical planning
(one example is Towers of Hanoi) but rather  common in the presence of non-determinism. An example of such a domain is  tireworld.\footnote{
  In  tireworld and variations, there are  roads leading to the goal with spare tires at some locations. 
  A drive action  moves the car from one location to the next and  may result in a flat tire. The fix action 
  requires  a spare at the location.}
  The number of states reachable by the solution policy is exponential in the length of the road as, while the car moves from each location to the goal,
it may leave a  spare behind    or not. Exponential policy size excludes all  (exact) FOND approaches except
symbolic methods like MBP and Gamer, and those using regression like PRP and Grendel. It is only for  some problems like tireworld, however, 
that PRP can compute correct policies using regression  without having to enumerate all  reachable  states.
This is achieved by a fast but incomplete verification algorithm \cite{prp}.  In general,  the correctness and the completeness of PRP
rely on this enumeration, and this means that   PRP, like methods that compute flat, non-compact policies, 
will not scale up in general  to problems with  policies that reach   an exponential number of states.\footnote{This limitation of PRP could be addressed potentially by using
 a complete verification algorithm working on the compact representation. This however would require regression over   non-deterministic
 actions \cite{rintanen:regression} and not just over deterministic plans.}

\medskip 

\noindent \textbf{Robust Non-Determinism.} In the first MDP planning competition, the planner that did best was a simple, classsical
replanner used \emph{on-line, } called  FF-replan \cite{ff-replan}. 
\Omit{
  discards the probabilistic information,  solves
the  determinism relaxation,  and applies  the resulting classical plan  to the  simulated MDP until observing a goal state or a
state $s$ not predicted  by the relaxation. In the latter case, the process is restarted  from $s$.  FF-replan won the  competition because it was  the only approach capable
of scaling  up to the size of the instances.
}
Little and Thiebaux argued then  that the   MDP and corresponding  FOND evaluation benchmarks
were not ``probabilistic interesting'' in general,  as they seldom  featured  \emph{avoidable deadends}, i.e., states with no weak plans which can  be avoided
in the way to the goal \cite{pip}. Avoidable deadends by themselves, however, present a challenge for \emph{incomplete, on-line} planners like 
FF-replan but not  for \emph{complete, off-line} classical replanners such  as PRP. The  reason  is that such planners
rely on  and require the ability to recover from  bad choices. Without the ability to ``backtrack''    in one way or the other, 
these planners wouldn't be complete. The  computational challenge for  complete  replanners  arises  not from the
presence of avoidable deadends but from  \emph{the number of ``backtracks''  required to find  a solution.}
In particular, FOND problems with avoidable deadends but  a small number of  weak  plans, impose no challenge
to complete replanners. \emph{There is indeed  no need to take non-determinism into account when reasoning about the ``future'' in complete replanners
 when the failure to do so  translates  into a small number of backtracks.}

The computational cost of reasoning about the future while ignoring non-determinism can be estimated.
For this, let $L_{\pi}(P)$ refer to the length of the shortest possible execution that
reaches the goal of $P$ from its initial state  following  a policy $\pi$ that solves $P$,
and let $L_{m}(P)$ be the minimum $L_{\pi}(P)$ over all such policies $\pi$. We refer
to the weak plans that have length smaller than $L_{m}$ as \emph{misleading plans.}
A misleading plan is thus a weak plan that does not  lead to a full policy, 
but which due to its length is likely to be found before weak plans that do.
As a result,  while non-classical approaches won't scale up to problems of large size,
and flat methods won't scale up to problems with  policies of exponential size, \emph{classical replanners will  tend to  fail 
on  problems that have an exponential number of misleading plans,} as the consideration of all such plans
is the price that they may have  to pay  for ignoring non-determinism when reasoning about the future.

We refer to the ability  to  handle  problems with an exponential number of misleading plans 
as \emph{robust non-determinism}.  Classical replanners like PRP  are  not bound to generate and discard each
of the misleading weak plans one by one given their  ability to propagate and generalize deadends. 
Yet, this component isn't  spelled out in sufficient detail in the case of PRP, and from the  observed
behavior (see below), this   is probably  done in  an  heuristic and limited manner.
Approaches that do not rely on classical planners but  which make use of  heuristics obtained
from deterministic relaxations are likely to face similar limitations.

While few existing benchmark domains give rise to an exponential number of  misleading  plans,
it is very  simple to come with variations of  them that do. Consider for example a version of  triangle tireworld
containing  two roads  to the goal: a short one of  length $L$  with spare  tires everywhere except in the last three locations, 
and a long road  of length $L' >> L$  with spare tires everywhere.  The car has capacity for a single spare,  but unlike the original domain
spares can  loaded   and  \emph{unloaded}.  In the instances $P$ of this  changed domain,
the number of  misleading plans grows \emph{exponentially}  in $L$. These are  weak plans  where the agent takes
the short road while moving spare tires around (for no good reason) in the way to the goal.
We will see  that in this revised domain,  a planner like Gamer does much better than PRP. The same
will be true for the  proposed SAT approach.

\section{SAT Approach to FOND Planning}

We provide a   SAT approach to FOND planning that is based on CNF encodings that are \emph{polynomial} in the number
of atoms and  actions. It  borrows elements  from both  the   SAT approach to classical planning
\cite{kautz:satplan} and previous SAT approaches to FOND and Goal POMDPs \cite{chitta:fond,chatterjee:sat}
that have   CNF encodings  that are polynomial in the number of states and hence \emph{exponential} in the
number of atoms. Our approach, on the other hand, relies on compact, polynomial encodings, 
and may result in compact policies too, i.e., policy representations that are polynomial
while  reaching  an exponential number of states. 

While the SAT approach to classical planning relies on atoms and actions that are indexed by time, bounded  by a given
horizon, the proposed SAT approach to FOND planning relies on atoms and actions indexed by controller states or nodes  $n$, whose
number is bounded by a given parameter $k$  that is  increased until a solution is found.
Each controller node $n$ stands for a partial state, and there are two special nodes: the initial
node $n_0$ where executions start,  and the goal node $n_G$ where  executions end.
The encoding only features deterministic actions $b$, so that non-deterministic actions $a=\{b_1, \ldots, b_n\}$
are encoded through the deterministic siblings   $b_i$. The atoms $(n,b)$ express that
$b$ is one of the (deterministic) actions to be applied in the controller node $n$, 
and constraints $(n,b) \rightarrow (n,b')$ and $(n,b) \rightarrow \neg (n,b'')$  express that all and only  siblings $b'$ of $b$
apply  in $n$ when $b$ applies. If $b$ is a deterministic action in the problem, it has no siblings. 
The atoms $(n,b,n')$ express that $b$ is applied in node $n$ and the control passes to node $n'$.
Below we will see how to get a strong cyclic policy from these atoms and how to execute it. 
%
%
For obtaining compact policies in this STRIPS non-deterministic setting where goals and action precondition are positive atoms
(no negation), we propagate negative information forward and positive information backwards. So, for example,
the encoding doesn't force $p$ to be true in $n'$ when $p$ is added by action $b$ and $(n,b,n')$ is true. Yet
if there are executions from $n'$ where $p$ is relevant and required, $p$ will be forced to be true in $n'$.
On the other hand, if $q$ is false in $n$ and not added by $b$, $q(n')$ is  forced to be false.

\subsection{Basic Encoding}

We present first the atoms and clauses of the CNF  formula  $C(P,k)$ for a FOND problem $P$
and a positive integer parameter  $k$ that provides the bound on the number of controller nodes.
Non-deterministic actions $a=\{b_1, \ldots, b_n\}$ in $P$ are encoded through the siblings   $b_i$.
For deterministic  actions $a$ in $P$, $a=\{b_1\}$.  The atoms in $C(P,k)$ are:

\begin{itemize} \denselist
	\item $p(n)$: atom $p$ true in controller state $n$,
	\item $(n, b)$: deterministic action $b$  applied in controller state $n$,
	\item $(n, b, n')$: $n'$ is next after applying  $b$ in  $n$,
	\item $ReachI(n)$:  there is path from  $n_0$ to $n$ in  policy,
	\item $ReachG(n, j)$: $\exists$ path from $n$ to $n_G$ with at most $j$ steps.
\end{itemize}

\noindent The number of atoms is quadratic in the number of controller states; this is different
than the number of atoms in the SAT encoding of classical planning that is linear in the horizon.
The clauses in $C(P,k)$ are given by the following formulas,  where   $P$ is given by a set of atoms,
the  set of  atoms true in the initial state $s_0$, a set of actions with preconditions and non-deterministic effects,
and the set of goals $G$:

\begin{enumerate}
 \item $\neg p(n_0)$ if $p \not\in s_0$ ; negative info in $s_0$
 \item $p(n_G)$ if $p \in G$ ; goal 
 \item $(n, b) \rightarrow p(n)$ if $p \in prec(b)$; preconditions
 \item $(n, b) \rightarrow (n, b')$ if $b$ and $b'$ are siblings 
 \item $(n, b) \rightarrow \neg (n, b')$  if $b$ and $b'$  not siblings 
 \item $(n, b) \iff \bigvee_{n'} (n, b, n')$; some next controller state
 \item $(n, b, n') \land \neg p(n) \rightarrow \neg p(n')$ if $p \not\in add(b)$; fwd prop.
 \item $(n, b, n') \rightarrow \neg p(n')$ if $p \!\in\! del(b)$; fwd prop. neg. info 
 \item $ReachI(n_0)$;  reachability from $n_0$  
 \item $(n, b, n') \land ReachI(n) \rightarrow ReachI(n')$
 \item $ReachG(n_G,j)$, $j=0, \ldots, k$, reach $n_G$ in $\leq$ $j$ steps
 \item $\neg ReachG(n,0)$ for all $n \not= n_G$ 
 \item $ReachG(n, j \!+\! 1) \iff  \bigvee_{b, n'} [(n, b, n') \land ReachG(n', j)]$
 \item $ReachG(n, j) \rightarrow ReachG(n, j \!+\! 1)$
 \item $ReachI(n) \rightarrow ReachG(n, k)$: if $n_0$ reaches $n$, $n$ reaches $n_G$.
\end{enumerate}

The control nodes $n$ form a labeled graph where the labels are the deterministic actions $b$,  $b \in a$,  for $a$ in $P$.
A control node $n$ represents a partial state comprised of the true atoms $p(n)$. Goals are true in $n_G$ and  preconditions of actions applied in $n$ are true
in $n$. Negative  information flows  forward along the edges, while positive information flows
backward,  so that  multiple system  states will be  associated with the same controller node in an
execution. The $ReachI$ clauses capture reachability from $n_0$, while $ReachG$ clauses capture reachability to $n_G$
in a bounded number of steps. The last clause states that any controller state $n$ reachable from $n_0$, must reach the goal node $n_G$.
Formula 13 is key for strong cyclic planning: it says that the goal is reachable from $n$ in at most  $j+1$ steps iff
the goal is reachable in at most $j$ steps from \emph{one} of its successors $n'$. For \emph{strong planning}, we will
change this formula so that the goal is reachable from $n$ in at most  $j+1$ steps iff the goal is reachable in at most $j$
steps from \emph{all}  successors $n'$.

For computing policies for a FOND problem $P$, a SAT-solver is called over  $C(P,k)$  where $k$ stands for the  number of
controller nodes $n$. Starting with  $k=2$ this bound is increased  by  $1$ until the formula is satisfiable. A solution policy 
can then be obtained from the  satisfying  truth assignment as indicated below.  If the formula $C(P,k)$
is unsatisfiable for $k=|S|+1$, then  $P$ has no strong cyclic solution.

\subsection{Policy}

A satisfying assignment $\sigma$ of the formula $C(P,k)$ defines a   policy  $\pi_{\sigma}$
that is a  function from controller states  $n$ into actions of $P$.
If the atom $(n,b,n')$ is  true in $\sigma$,  $\pi_{\sigma}(n)=b$ if $b$ is a deterministic action in $P$
and $\pi_{\sigma}(n)=a$ if $b \in a$ for a non-deterministic action $a$ in $P$.

\Omit{
For this,   consider the  joint states $\pair{n,s}$
given by the controller and system states $n$ and $s$. Initially  this pair is $\pair{n_0,s_0}$, and
and after doing the  action $a=\pi_{\sigma}(n)$ from $P$ in the joint state $\pair{n,s}$ and observing the state $s'$,
$s' \in F(a,s)$, the next pair is $\pair{n',s'}$ where $n'$ is the unique controller state such that $(n,a,n')$ is true in $\sigma$
when $a$ is deterministic, or the unique controller state such that $(n,b,n')$ is true in $\sigma$, $b \in a$,
and $s'$ is the state that deterministically follows $b$ in $s$.\footnote{This last controller is unique as 
siblings $b'$ of $b$ that are also part of atoms $(n,b',n')$ that must be true, must differ from $b$ in some STRIPS
effect, and hence cannot map $s$ into $s'$.}
}

For applying the \emph{compact  policy}  $\pi_{\sigma}$, however, it is necessary  to keep track of the controller state.
For this, it is convenient to consider a second  policy $\pi'_{\sigma}$ determined by $\sigma$,
this one being  a standard  mapping of   states into actions  over an  \emph{extended  FOND} $P_{\sigma}$
that denotes  a FOND   model $M_{\sigma}$. In this (cross-product)  model,  the  states are
pairs   $\pair{n,s}$ of  controller and a system states, the initial state  is   $\pair{n_0,s_0}$, the goal states
are  $\pair{n_G,s}$ for $s \in  S$,  and the set  $A_{\sigma}(\pair{n,s})$ of actions applicable in $\pair{n,s}$
is restricted to the \emph{singleton set}  containing  the action $a= \pi_{\sigma}(n)$ for the compact policy  $\pi_{\gamma}$ above.
The transition function $F_{\sigma}(a,\pair{n,s})$  results  in  the pairs $\pair{n',s'}$ where $s' \in F(a,s)$ and $n'$ is the
unique controller state for which  a)~the atom  $(n,a,n')$ is true in $\sigma$ when $a$ is deterministic, or b)~the atom  $(n,b,n')$  is true in $\sigma$ for
$b \in a$ with $s'$  being the unique successor of $b$ in $s$ otherwise.

In the extended FOND $P_{\sigma}$ there is a just one policy, denoted as  $\pi'_{\sigma}$, that over  the reachable pairs $\pair{n,s}$
selects  the only  applicable action  $\pi_{\sigma}(n)$. We say that the \emph{compact policy} $\pi_{\sigma}$ is a strong cyclic (resp. strong)
policy for $P$ iff $\pi'_{\sigma}$ is a strong cyclic (resp. strong) policy for $P_{\sigma}$.

\subsection{Properties}

We show that the SAT approach is sound and complete for strong cyclic planning. We  consider  strong planning later. 

\begin{theorem}[Soundness]
If $\sigma$ is a satisfying assignment for $C(P,k)$, the compact policy $\pi_{\sigma}$ is a  strongly cyclic solution for $P$.
\end{theorem}

{\footnotesize
\begin{sketch} 
  Let $\sigma$ be a satisfying assignment  for $C(P,k)$ and let $\pi'_{\sigma}$ be the only policy  for the  FOND $P_{\sigma}$ above.
  We need to show that $\pi'_{\sigma}$ is a strong cyclic policy for $P_{\sigma}$. First, it can be shown inductively that if $\pair{n,s}$ is reachable
  by $\pi'_{\sigma}$ and  $p(n)$ is true in $\sigma$,  $p$ is true in $s$.
  \Omit{
    By induction, if the pair is reachable in $0$ steps,
  the pair must be $\pair{n_0,s_0}$ but then $p(n_0)$ implies $p$ true in $s_0$,  as otherwise,  clause 1 of the encoding would
  make $p(n_0)$ false. If this is true  for executions of length up to $r > 0$  and that the last transition in an execution of length
  $r + 1$ is $(n, s), \, b, \, (n', s')$ where $b$ is one of the outcomes of the action $a$ prescribed for the controller state $n$. We want to show that if $p(n')$ is true in $\Sigma$, $p$ must be true in $s'$. Let's assume otherwise; namely that $p(n')$ is true in $\Sigma$ but $p$ is false in $s'$. There are two cases to consider.If $p \not\in Add(b)$, since $p$ is false in $s'$, it must be false in $s$.
  From the induction, this means that $p(n)$ must be false in $\Sigma$. Yet this is not possible, as formula 7 of the encoding ensures that $p(n')$ should then be false as well, in contradiction with the assumption. n
  If $p \in Add(b)$, $p$ must be true in $s'$, which also contradicts the assumption. As a result, $p(n')$ implies that $p$ must be true in $s'$.
  }
This also implies that if the extended state $\pair{n,s}$ is reachable in $P_{\sigma}$ and $\pi_{\sigma}(n)=a$,  then for  each precondition $p$ of $a$,
 $p$ is true in $s$. If the policy reaches a joint state $(n, s)$, i.e. there is a path $(n_0, s_0)$, $(n_1, s_1)$, $..., (n, s)$, $ReachI(n)$ will be set to true by clause 10, thus forcing $ReachG(n, k)$ to be true (clause 15). In order to satisfy  13, there must be a path $(n_1, s_1), (n_2, s_2), ..., (n_j, s_j)$ of at most $k$ steps, with $n_1 = n$ and $s_1 = s$, such that $n_j$ is $n_G$ and $s_j$ is a goal state. This follows from clause 2. 
\end{sketch}
}

\medskip

For showing completeness, if $\pi$ is  a strong cyclic  policy  for  $P$, let us define the \emph{relevant atoms} in a state $s$
reachable by $\pi$ from $s_0$ as the atoms $p$ that are true in $s$ such that there is an execution $\tau$  from $s$ to a state $s'$ such that $p$
is not deleted from $s$ to $s'$ and either 1)~$s'$ is a goal state and $p$ is a goal atom, or 2)~$s'$ is not a goal
state and $p$ is a precondition of the action $\pi(s')$. For a reachable state $s$, let $\hat{s}$ be the reachable partial state
comprising the atoms in $s$ that are relevant given $\pi$. We call these  partial states  the \emph{$\pi$-reduced states.}
Then, completeness can be expressed as follows:

\begin{theorem}[Completeness]
  Let $\pi$ be a strong cyclic policy for $P$ and let $N_{\pi}(P)$ represent  the number of different $\pi$-reduced states.
  Then if $k \ge N_{\pi}(P)$,  there is an assignment $\sigma$ that satisfies $C(P,k)$ and $\pi_{\sigma}$ is a compact strong cyclic policy
   for $P$.
  \label{thm:compl}
\end{theorem}

{\footnotesize
  \begin{sketch}
     Consider the  problem $P_{\pi}$ whose states are the  pairs $\pair{r,s}$, where $r$ is the set of  relevant atoms  in $s$ given
     the policy $\pi$, which represents  the original problem $P$  but with the states $s$ augmented with the set of atoms $r$
     determined by $s$ and $\pi$.      The policy $\pi_A(\pair{r,s}) = \pi(s)$ is clearly a strong cyclic solution to $P_{\pi}$, and it can be
      shown    that a compact  policy $\pi'$ can be defined over the ``reduced states'' $r$ only,  such that
     $\pi_B(\pair{r,s}) = \pi'(r)$ is also a strong cyclic solution to $P_{\pi}$. For this, if $r$
     is the set of relevant atoms in  $s$,  it suffices to set $\pi'(r)$ to $\pi(s)$ for any such
     $s$.   It needs  then to be shown   that there is a truth assignment $\sigma$ that satisfies $C(P,k)$  for $k$ in the theorem,
      where the    controller    nodes $n_r$ are associated with  the ``reduced states'' $r$,  $p(n_r)$ is true  in $\sigma$ iff $p \in r$,
      and $(n_r,b,n_{r'})$ is true in $\sigma$ iff $F_{\pi}(b,\pair{r,s})= \{\pair{r',s'}\}$ for some states $s$ and $s'$. 
      \end{sketch}
}

\medskip

\noindent Finally, the  policy $\pi_{\sigma}$ is   compact in the following sense:

\begin{theorem}[Compactness]
The size of the  policy $\pi_{\sigma}$ for a truth  assignment $\sigma$ satisfying $C(P,k)$
can be exponentially smaller than the number of states reachable by $\pi_{\sigma}$.
\end{theorem}


{\footnotesize
    \begin{sketch}
   A single example suffices to show that the number of  states reachable with the policy $\pi_{\sigma}$
    can be exponentially larger than the size of the policy. For this, 
    consider  a version of tireworld $P$  where there is a single road to the goal
      with locations $L_0, \ldots, L_m$  where $L_m$ is the goal and there is a spare tire in each location.
      The \emph{number of states} reachable by the solution  policy $\pi$ is \emph{exponential} in $m$ as the goal may be reached leaving
      behind any 0/1 distribution of spares over the locations $L_1, \ldots, L_{m-1}$. However,
      when the execution of the  policy $\pi$ reaches a state $s_i$ where the car is at location $L_i$, only the atoms $spare(L_j)$
      with $j \ge i$ are relevant in $s_i$,  and these are atoms are then  all true,   with the possible  exception of $spare(L_i)$. 
      The atoms $spare(L_j)$ for $j < i$, which may  be true or false, are  not relevant then.
      As a result, the \emph{number of reachable}  $\pi$-reduced states,   unlike the number  reachable states,
      is $2m$. Theorem~\ref{thm:compl} implies   that for  $k =  2m$,  there must be  an assignment
       $\sigma$ that satisfies $C(P,k)$, and hence  a  compact strong cyclic policy  $\pi_{\sigma}$  for $P$
       with $2m$ controller states that reaches a number of system states that is exponential in  $m$. 
      \end{sketch}
}

\section{Optimizations}

We introduced simple  extensions and modifications  to the SAT encoding   to make it more efficient and scalable
while maintaining its formal properties. The  actual encodings used in the experiments   feature extra variables  $(n, n')$
that are  true iff $(n,b,n')$ is true for some action $b$.  Also, since the  number of variables  $(n, b, n')$ grows quadratically
with the number of control nodes, we substitute  them by variables $(n, B, n')$ where $B$ is the action name  for action $b$
without the arguments. It is assumed that siblings $b$ and $b'$ of  non-deterministic actions  $a$ get different action names
by the parser. As a result, the conjunction $(n,B,n') \land (n,b)$ can be used in substitution of $(n,b,n')$.
Similarly,  add lists of actions tend to be short, resulting in a huge   number of clauses of type 7
for capturing forward propagation of negative information. These clauses are replaced by 

\begin{enumerate} 
\item[] 7'. $(n, n') \land \neg p(n) \rightarrow \neg p(n') \lor \bigvee_{b: p \in add(b)} (n, b)$
\item[] 7''. $(n, B, n') \land (n, b) \land \neg p(n) \rightarrow \neg p(n')$, 
\end{enumerate}
the last  clause for actions $b$ that do not add $p$ but have siblings that do only.
Finally,  extra formulas are added for  breaking   symmetries that result from
exchanges in the names (numbers) associated with different control nodes, other than   $n_0$ and $n_G$, that
result in  equivalent controllers.

\Omit{
\begin{itemize}
\item $(n_j, n_k) \rightarrow \bigvee\limits_{i \leq j} (n_i, n_{k-1})$
\item $(n_i, n_k) \iff (n_j, n_i) \land \bigwedge\limits_{k<j} \neg (n_k, n_i)$
\item $(n_i,n_j) \land (n_i, n_k) \land (n_j,n_i) \land (n_k,n_i) \land (n_i, a) \rightarrow (n_i, a_0, n_j) \land (n_i, a_1, n_{j+1})$ 
\end{itemize}
We tried more sophisticated and systematic encodings \cite{sym-breaking} but they didn't result in uniform improvements
of performance. The second and third formulas are for non-deterministic actions with two possible outcomes only. *** Not clear; recheck, explain ***
*** Removed $L$  $L(n,n')$??? ****
}

\begin{table*}[h]
\centering
\small
\begin{tabular}{|l|c|c||c|c|c||c|c|c||c|c|c||c|c|}
\hline
\multicolumn{3}{|c||}{} & \multicolumn{3}{|c||}{SAT approach} & \multicolumn{3}{|c||}{PRP} & \multicolumn{3}{|c||}{MYND} & \multicolumn{2}{|c|}{GAMER} \\ \hline
\footnotesize{Domain (\# inst)} & \footnotesize{\#at} & \footnotesize{\#acts} & \footnotesize{\%solve} & \footnotesize{time} & \footnotesize{size} & \footnotesize{\%solve} & \footnotesize{time} & \footnotesize{size} & \footnotesize{\%sol} & \footnotesize{time} & \footnotesize{size} & \footnotesize{\%solve} & \footnotesize{time} \\ \hline \hline

acrobatics (8) & 67 & 623 & 50 & 572.5 & 17 & \textbf{100} & 20.1 & 127 & \textbf{100} & 4.5 & 126 & 87 & 5.2 \\ \hline
beam walk (11) & 746 & 2231 & 27 & 43.2 & 20 & \textbf{100} & 27.6 & 1488 & \textbf{100} & 126.6 & 1487 & 90 & 41.2 \\ \hline
faults (20) & 43 & 35 & \textbf{100} & 7.2 & 12 & \textbf{100} & 0.1 & 9 & \textbf{100} & 46.7 & 45 & \textbf{100} & 2.5 \\ \hline
faults (20) & 84 & 92 & 65 & 684.4 & 19 & \textbf{100} & 0.2 & 16 & \textbf{100} & 69.9 & 261 & 70 & 128.8 \\ \hline
faults (15) & 129 & 173 & $0^T$ & - & - & \textbf{100} & 0.2 & 20 & \textbf{100} & 59.3 & 1258 & 20 & 152.9 \\ \hline
first resp (30) & 28 & 41 & 63 & 129.6 & 13 & \textbf{66} & 0.1 & 13 & 63 & 71.8 & 11 & 50 & 3.5 \\ \hline
first resp (40) & 99 & 436 & 57 & 194.2 & 13 & \textbf{82} & 1.6 & 17 & 77 & 240.2 & 18 & 17 & 1.2 \\ \hline
first resp (30) & 172 & 1333 & 30 & 174.4 & 12 & \textbf{76} & 0.3 & 24 & 36 & 433.4 & 17 & 3 & 1.0 \\ \hline
t. tireworld (20) & 669 & 1406 & 15 & 149.6 & 18 & \textbf{100} & 5.7 & 125 & 35 & 136.3 & 9382 & 30 & 1.7 \\ \hline
t. tireworld (20) & 4129 & 9006 & $0^T$ & - & - & \textbf{100} & 374.6 & 365 & $0^M$ & - & - & $0^M$ & - \\ \hline
zenotravel (15) & 377 & 8424 & 33 & 243.9 & 15 & 	\textbf{100} & 5.1 & 54 & $0^P$ & - & - & 6 & 0.0 \\ \hline
elevators (10) & 64 & 58 & 70 & 28.8 & 18 & \textbf{100} & 0.2 & 29 & \textbf{100} & 71.1 & 28 & \textbf{100} & 4.4 \\ \hline
elevators (5) & 123 & 116 & $0^T$ & - & - & \textbf{100} & 1.1 & 81 & 60 & 2221.9 & 91 & 20 & 464.9 \\ \hline
blocks (15) & 78 & 1350 & 66 & 27.6 & 13 & \textbf{100} & 0.2 & 15 & 93 & 33.2 & 16 & 66 & 34.8 \\ \hline
blocks  (15) & 238 & 8116 & $0^T$ & - & - & \textbf{100} & 0.9 & 33 & 53 & 927.3 & 40 & $0^M$ & - \\ \hline
tireworld (15) & 63 & 304 & \textbf{80} & 6.5 & 7 & \textbf{80} & 0.1 & 10 & \textbf{80} & 11.6 & 7 & 73 & 126.5 \\ \hline
earth obs (20) & 46 & 87 & 40 & 697.1 & 16 & \textbf{100} & 0.4 & 62 & \textbf{100} & 192.4 & 73 & $0^P$ & - \\ \hline
earth obs (20) & 110 & 224 & 5 & 3510 & 38 & \textbf{100} & 1.8 & 234 & 50 & 459.4 & 138 & $0^P$ & - \\ \hline
\hline 
miner (30) & 587 & 1209 & \textbf{100} & 160.6 & 21 & 26 & 556.5 & 19 & $0^T$ & - & - & $0^P$ & - \\ \hline
miner (21) & 1410 & 2920 & \textbf{100} & 1102 & 25 &	6 & 721.3 & 25 & $0^T$ & - & - & $0^P$ & - \\ \hline
ttire spiky (11) & 256 & 484 & \textbf{90} & 911.0 & 26 & $0^T$ & - & - & $0^T$  & - & - & 18 & 115.8 \\ \hline
doors (15) & 48 & 69 & 93 & 597.0 & 20 & 80 & 3.2 & 22 & 73 & 288.3 & 1486 & \textbf{100} & 4.2 \\ \hline
islands (30) & 100 & 333 & \textbf{100} & 8.1 & 8 & 76 & 167 & 5 & 30 & 127.1 & 4 & 43 & 1.9 \\ \hline
islands (30) & 388 & 1588 & \textbf{96} & 496.4 & 12 & 26 & 256.7 & 11 & 13 & 85.3 & 10 & 16 & 2.8 \\ \hline
ttire  truck (24) & 61 & 107 & \textbf{100} & 6 & 14 & 37 & 185.1 & 18 & 33 & 73.8 & 12 & $0^P$ & - \\ \hline 
ttire truck (25) & 80 & 150 & \textbf{100} & 96.8 & 19 & 32 & 500 & 27 & 16 & 860.5 & 17 & $0^P$ & - \\ \hline
ttire truck (25) & 101 & 198 & \textbf{88} & 193.8 & 19 & 16 & 384.5 & 21 & 8 & 24 & 17 & $0^P$ & - \\ \hline
\end{tabular}
\caption{Results for strong cyclic planning.   Each line contains the domain name, number of instances in parenthesis, then avg. number
  of atoms and actions per instance followed by \% of instances solved, avg. time in seconds, and avg. policy size for each of the four planners.
  Domains that involve many instances of very  different sizes are split in multiple lines. New domains in the bottom part. Coverage expressed by percentages as
  different number of instances per line. Best coverages in bold. When coverage is $0$ we write $0^T$, $0^M$, or $0^P$ to indicate
  if the problem is a time out, a memory out, or a parsing error.}
\label{table-results}
\end{table*}

\section{Experimental Results}

We have compared our SAT-based FOND solver with some of the best existing planners; namely, PRP, MyND, and Gamer.\footnote{
 The version of PRP  is the newest,  8/2017, from \url{https://bitbucket.org/haz/deadend-and-strengthening}.
  MyND was obtained from \url{https://bitbucket.org/robertmattmueller/mynd}, while we managed to  get Gamer only from the authors of MyND.}
The four planners were run on an AMD Opteron 6300@2.4Ghz, with time and memory limits of 1h and 4GB (10GB for Gamer).
  The SAT solver used was MiniSAT \cite{minisat}. We used  the  FOND domains and  instances available from   previous
  publications, and added  new domains of  our own. We explain them briefly below.

  \noindent \textbf{Tireword Spiky:} A  modification  of triangle tireworld.  The main difference is that  the agent (car)  can  drop spare tires,
  not just pick them up, while  holding one spare at a time  at most. In addition,  not all roads can produce a flat tire
   (i.e., there are normal and  spiky roads).
  In these  instances  there are two roads to the goal, one shorter with two spiky segments,   one after the other,  and not enough spares, 
  and a longer path with one spiky segment only. On the first location of the short path there are  several spare tires.
  The misleading plans take  the short road to the goal,   moving  spares around with no purpose.
  
  \noindent \textbf{Tireworld Truck:} A modification  of Tireworld Spiky where there are  a few  spiky  segments.
    In  this version, all the spares are in the initial location and there is  a truck there too that can load  and unload tires,
   and is not affected by spiky  roads. The truck and the car cannot be in the same location except  for the initial location.
   The solution  is for the truck to pick up the spares that the car will need and place them at their proper places, returning to the initial
   location, before  the car leaves.

 \noindent \textbf{Islands.} Two grid-like islands of size $n \times n$ each are connected by a bridge.
 Initially the agent is in island 1 and the goal is to reach a specific location in island 2.
 There are two ways of doing this: the short way is to  swim  from island 1 to 2, and the long  way is to go  to the bridge and cross it.
 Swimming is  non-deterministic as the agent may drown.  Crossing the bridge is possible when  the bridge is free, else the animals that block it
 have to be moved away first. The misleading plans are those  where the agent   moves  some animals away  and then swims to the other island. 

 \noindent \textbf{Doors:} A row of  $n$ rooms one after the other connected through  doors. The  agent has to move  from the first  to the last.
   Every time the agent enters a room,  the in and out  doors of the room  (except for  first and last rooms) open or close non-deterministically.
   There are actions for entering a room when the  door is open and when the door is closed, except for the last room that requires a key when closed. 
   The key is initially in the first room. The agent cannot move backwards. The solution is to pick up  the key first and  then head to the end room.
   A version of this domain was considered in \cite{cimatti:three-models}. 

    \noindent \textbf{Miner.} An agent has to  retrieve a number of items that can be found in two regions. In each region, an  item can be digged out
    by moving stones. In the places that are closer to the agent  these operations are not fully safe. The misleading plans
     are those where the items are sought at the close but unsafe sites, possibly moving stones around.

     \Omit{
  The agent is placed on a grid, and has to collect some items while avoiding fake ones. Some fof the items are fake and might have negative effects (non deterministic), while some other are real, but are hidden. In order to find those, the agent must use a rock to press a botton. \textbf{Basicamente hay piedras que se pueden mover, y hay que moverlas para poder acceder a los items buenos. Los items malos pueden matarte. Digamos, piedras tipo tires y el deadend de morir siguiendo el camino facil y corto. Por otro lado, los items buenos estan mas alejados que los malos. Lo pongo para dejar todo completo, pero feel free de sacarlo de aca y la tabla}
   }

   \Omit{
   \noindent \textbf{Tireworld Used:} This is exactly the  triangle tireworld with three  changes: a location with a spare has an infinite supplies of spares
   (atom not deleted after use) but    using a spare tire in a location flags  the presence of one or more flat tires    to be collected
   (not part as part of the goal though). Last,  when a car moves from one location to another, it may not just experience a flat tire,
   a third non-deterministic outcome of the action may teleport the car back 
     to the initial location. The solution policy though  is the same as for the original problem.
   }

   \Omit{
   \noindent \textbf{Spiky-1.} Simplified version of tireworld spiky. The agent can pick spares but cannot drop them in other locations. There are two paths, one that leads nowhere but has infinite spares, and one that leads to the goal, has no spares, and a spiky road between two locations.
. \textbf{Spiky-2.} Like Spiky-1, but the path with infinite spares leads to the goal with 3 consecutive spiky roads (which makes it unusable when searching or strong cyclic solutions).
   }

   \begin{table*}[]
\centering
\small
\begin{tabular}{|l|c|c||c|c|c||c|c|c||c|c|}
\hline
\multicolumn{3}{|c||}{} & \multicolumn{3}{|c||}{SAT approach} & \multicolumn{3}{|c||}{MYND} & \multicolumn{2}{|c|}{GAMER} \\ \hline
\footnotesize{Domain (\# instances)} & \footnotesize{\#atoms} & \footnotesize{\#acts} & \footnotesize{\%solve} & \footnotesize{time} & \footnotesize{size} & \footnotesize{\%solve} & \footnotesize{time} & \footnotesize{size} & \footnotesize{\%solve} & \footnotesize{time} \\ \hline \hline
zenotravel (15)         & 377  & 8424  & \textbf{33}  & 130.7 & 15   & $0^P$         & -      & -    & 6           & 0.0  \\ \hline
elevators (10)          & 64   & 58    & 70           & 14.5  & 18   & \textbf{80} & 9.3    & 19   & \textbf{80}  & 6.5  \\ \hline
elevators (5)           & 123  & 116   & $0^T$          & -     & -    & \textbf{20} & 34.0   & 66   & $0^{M,T}$           & -    \\ \hline
miner (30)              & 587  & 1209  & \textbf{100} & 90.9  & 21   & $0^T$         & -      & -    & $0^P$           & -    \\ \hline
miner (21)              & 1410 & 2920  & \textbf{100} & 446.5 & 25   & $0^T$         & -      & -    & $0^P$           & -    \\ \hline
tireworld-spiky (11)    & 256  & 484   & \textbf{90}  & 225.7 & 26   & 63          & 1149.5 & 27   & 18           & 99.8 \\ \hline
doors (15)              & 48   & 69    & 86           & 160.6 & 20   & 73          & 166.4  & 1486 & \textbf{100} & 4.3  \\ \hline
tireworld (15)          & 63   & 304   & \textbf{80}  & 4.1   & 7    & 20          & 8.0    & 1    & 20           & 0.0  \\ \hline
islands (30)            & 100  & 333   & \textbf{100} & 7.7   & 8    & 96          & 97.7   & 4    & 43          & 2.4  \\ \hline
islands (30)            & 388  & 1588  & \textbf{100} & 334.4 & 12   & 26          & 338.8  & 10   & 20           & 5.1  \\ \hline
triangle-tireworld (20) & 669  & 1406  & 15           & 112.3 & 18   & 15          & 90.3   & 2182 & \textbf{30}  & 1.7  \\ \hline
\end{tabular}
\caption{Results for strong planning over  domains with strong solutions in  Table 1}
\label{table-results-strong}
\end{table*}

The results for the four planners over the existing and new domains are shown in Table~\ref{table-results}.
Domains that involve many instances of widely  different sizes are split into  multiple lines,  and coverage is  expressed
by percentages as   different lines involve different  number of instances. The best coverages for each line are shown in
bold.  Overall,  PRP does best, yet in order to understand the strength and limitations of the various planners,
it is useful to consider the problem size, policy size, and type of non-determinism, and also whether the problems are old
or new. Indeed, it turns out that PRP is best among the existing domains, most of which predate PRP. On the other hand,
for the new domains, the SAT approach is best. Indeed, PRP can deal with very large problems (as measured by the number of atoms
and actions), and can also produce large controllers, with hundreds and even thousands of partial or full states.
To some extent, MyND is also pretty  robust to problem and controller size, but does not achieve the same results.
On other hand, the SAT approach has difficulties scaling up to problems that require controllers with more than
30 states, in particular, if  problem size is large too. In classical planning, the SAT approach has a similar
limitation with  long sequential plans.  In our SAT approach to FOND, this limitation is compounded by the fact that the CNF encodings are
quadratic in the number of controller states.  On the other hand, the table shows that the SAT approach is the most robust
for dealing with problems with many misleading  plans, as in several of the new domains,
where  not taking non-determinism into account when reasoning about the future
makes the ``optimistic'' search for plans computationally unfeasible. 

\subsection{Discussion}

Overall,  PRP scales up best to problem size and even policy size, but it doesn't scale up as well
as the SAT approach on  forms of non-determinism that involve many misleading  plans. 
Clearly, PRP has excellent  coverage on a domain like Triangle Tireworld that also involve
an exponential number of misleading plans, but this  is achieved by  methods for identifying,
generalizing, and propagating deadends that are not general.
The SAT approach does better in handling richer forms of non-determinism because of its ability to reason
in parallel about the different, branching futures  arising   from non-deterministic actions.
The challenge for the  SAT approach is to scale up more robustly
to problem size, and in particular, to controller size. For classical planning, similar challenges have been addressed quite successfully
through a number of techniques, including better encodings, different forms of parallelism, variable
planning-specific selection heuristics, and alternative ways for increasing the time horizon \cite{rintanen:h}.
For SAT approaches to FOND, this is all to be explored, and additional techniques like incremental SAT solving
should be explored as well \cite{incremental-sat-solving,inc-sat-planning}.

\section{Strong Planning}

The SAT encoding above is for computing strong cyclic policies. For computing strong policies instead,
the formula 13 in the encoding has to be replaced by 13':

\begin{enumerate}
  \setcounter{enumi}{13}
\finalitem $ReachG(n, j \!+\! 1) \!\! \iff \!\!  \bigwedge_{b, n'} [(n, b, n') \!\rightarrow\! ReachG(n', j)]$
\end{enumerate}

\noindent meaning that for a node $n$ to be at less than  $j+1$ steps from the goal, all its successors
must be at less than $j$ steps from the goal.
Table~\ref{table-results-strong}  shows  the figures   for the resulting SAT-based strong FOND planner in comparison with MyND
and Gamer used in strong planning mode. The domains are those from  Table 1 where at least  one of the strong  planners found a
solution.  In this case, the results over the existing domains are mixed, with the SAT approach doing best in one of the
domains, and MyND and Gamer doing best on the other two. The SAT approach is best on the new domains with the exception 
of Doors where  Gamer does better.

\section{Strong and Cyclic Planning Combined}

A feature of the SAT approach that is not shared by either classical replanners, OBDD-planners, or
explicit AND/OR search  approaches like MyND and Grendel, is that in SAT, it is very simple to
reason with a combination of actions that can be assumed to be fair, with actions that cannot,
leading to a form of planning that is neither strong nor strong cyclic. We call this
Dual FOND planning.\footnote{Related issues are discussed in \cite{camacho:dual}.}

Dual FOND planing is planning with  a FOND problem $P$ where some of the actions are tagged as
fair, and the others  unfair. For example, consider  a problem featuring   a planning agent and an  adversary,
one in front of the  other in the middle row of  a   $3 \times 2$ grid (two columns): the agent on the left, the adversary on the right,
and the agent must reach a position on the right. The agent  can move up and down non-deterministically, moving 0, 1, or 2 cells,
without ever leaving the grid, he  can also  wait, or he can  move  to the opposing cell on  the right 
if that position is empty.  Every turn however, the adversary  moves 0 or 1 cells, up or down.
The solution to the problem is for the agent to keep moving up and down until he is  at vertical distance of
2 to the opponent, then moving right. This strategy is not a strong or a strong cyclic policy,
but  a \emph{dual policy}. 

A state trajectory $\tau$  is \emph{fair} for   a Dual FOND problem $P$ and a policy $\pi$  when infinite occurrences of a state
$s$ in $\tau$, \emph{where $a=\pi(s)$ is a fair action}, implies infinite occurrences of transitions $s,s'$ in $\tau$
for each successor $s' \in F(a,s)$. A solution to a Dual FOND problem $P$ is a policy $\pi$ such that all
the fair trajectories induced by $\pi$ are goal reaching.  Strong cyclic  and strong planning are special cases of
Dual FOND planning when all or none of the actions are fair.  A sound and complete SAT formulation of Dual FOND planning
is obtained by introducing the atoms $(n,fair)$, that will be true if the action chosen in $n$ is fair, 

\begin{enumerate}
\item[] 16. $(n, fair) \iff \bigvee_{b} (n, b)$, $b$ among fair actions 
\item[] 17. $\neg (n, fair) \iff \bigvee_{b} (n, b)$, $b$ among unfair actions
\end{enumerate}

\noindent and replacing 13 and 13'  by:

\begin{enumerate}
\item[] 13''.  $[(n,fair) \rightarrow \textrm{13}] \, \land \, [\neg (n,fair) \rightarrow \textrm{13'}]$
\end{enumerate}
\noindent  where 13 and 13' are the formulas above for strong cylic and strong planning.
The above encoding captures dual FOND planning in the same way that
the first encoding captures strong cyclic planning.

We have run  some experiments for dual planning too, for  the example above where the  two agents
move over a $n \times 2$ grid. We tried values of $n$ up $10$, and the resulting
dual policy is the one mentioned above, where the agent keeps moving up and down
until leaving the adversary behind. Notice that strong, strong cyclic, and dual FOND
planning result from simple changes in some of the clauses. This flexibility is a strength of the
SAT approach that is  not available in  other approaches that require different algorithms
in each case. 

\section{Conclusions}

We have introduced the first SAT formulation for FOND planning that is compact
and can produce compact policies. Small changes in the formulation account for
strong, strong cyclic, and a combined  form  of strong and strong cyclic planning,
that we call dual FOND planning, where some actions are assumed fair and the others
unfair. From a computational point of view, the  SAT approach performs well 
in  problems that are not too large and that do not require  large controllers,
where  it is  not  affected by the presence of a large number of misleading  plans.
Classical replanners like  PRP and explicit AND/OR search planners like MyND can
scale up to larger problems or problems with larger controllers respectively,
but do not appear to be  as robust to  non-determinism.

\medskip

\subsection*{Acknowledgments}
We thank Miquel Ram\'{\i}rez and Jussi Rintanen for useful exchanges,
Chris Muise for answering  questions about PRP, and Robert Matm\"{u}ller
for the code of both MyND and Gamer. H. Geffner is partially funded by  grant TIN2015-67959-P from MINECO, Spain.


\bibliography{control}
\bibliographystyle{aaai}

\end{document}